%% file: main.tex
\documentclass[runningheads]{llncs}
\usepackage{wrapfig}
\usepackage{multirow}

\usepackage{marvosym}
\usepackage{eccv}

\usepackage{eccvabbrv}

\usepackage{graphicx}
\usepackage{booktabs}

\usepackage[accsupp]{axessibility}  % Improves PDF readability for those with disabilities.

\usepackage{hyperref}

\usepackage{orcidlink}

\begin{document}

% ---------------------------------------------------------------
% TODO REVIEW: Replace with your title
\title{DiffFAS: Face Anti-Spoofing via Generative Diffusion Models} 
% TODO REVIEW: If the paper title is too long for the running head, you can set
% an abbreviated paper title here. If not, comment out.
\titlerunning{DiffFAS}

% TODO FINAL: Replace with your author list. 
% Include the authors' OCRID for the camera-ready version, if at all possible.
\author{
Xinxu Ge\inst{1}\orcidlink{0009-0008-2622-4631} \and
Xin Liu\inst{2}\thanks{Corresponding authors.}\orcidlink{0000-0002-2242-6139} \and
Zitong Yu\inst{3}$^{\star}$\orcidlink{0000-0003-0422-6616} \and
Jingang Shi\inst{4} \and
Chun Qi\inst{4} \and
Jie Li\inst{4} \and
Heikki Kälviäinen\inst{2}\orcidlink{0000-0002-0790-6847}
}

% TODO FINAL: Replace with an abbreviated list of authors.
\authorrunning{X. GE et al.}
% First names are abbreviated in the running head.
% If there are more than two authors, 'et al.' is used.

% TODO FINAL: Replace with your institution list.
\institute{Tianjin University, China\and
Lappeenranta-Lahti University of Technology LUT, Finland\and
Great Bay University, China \and
Xi'an Jiaotong University, China \\
\email{xinxu\_ge928@tju.edu.cn}, \email{linuxsino@gmail.com},  \email{yuzitong@gbu.edu.cn},\\ \email{\{jingang, qichun, jielixjtu\}@xjtu.edu.cn}, \email{heikki.kalviainen@lut.fi}
}

\maketitle

\input{sec/0_abstract}    
\input{sec/1_intro}
\input{sec/2_Related_work}

\input{sec/3_method}

\input{sec/4_experiment}

\input{sec/5_ablation}

\input{sec/6_conclusion}

\section*{Acknowledgments}
This work was supported in part by the National Natural Science Foundation of China under Grant 62171309 and 62306061, Guangdong Basic and Applied Basic Research Foundation (Grant No. 2023A1515140037), Open Fund of National Engineering Laboratory for Big Data System Computing Technology (Grant No. SZU-BDSC-OF2024-02).

\bibliographystyle{splncs04}
\bibliography{main}

\input{sec/X_suppl}

% \bibliographystyle{splncs04}
% \bibliography{main}
\end{document}

%% file: sec/0_abstract.tex
\begin{abstract}
Face anti-spoofing (FAS) plays a vital role in preventing face recognition (FR) systems from presentation attacks. Nowadays, FAS systems face the challenge of domain shift, impacting the generalization performance of existing FAS methods. In this paper, we rethink about the inherence of domain shift and deconstruct it into two factors: image style and image quality. Quality influences the purity of the presentation of spoof information, while style affects the manner in which spoof information is presented. Based on our analysis, we propose DiffFAS framework, which quantifies quality as prior information input into the network to counter image quality shift, and performs diffusion-based high-fidelity cross-domain and cross-attack types generation to counter image style shift. DiffFAS transforms easily collectible live faces into high-fidelity attack faces with precise labels while maintaining consistency between live and spoof face identities, which can also alleviate the scarcity of labeled data with novel type attacks faced by nowadays FAS system. We demonstrate the effectiveness of our framework on challenging cross-domain and cross-attack FAS datasets, achieving the state-of-the-art performance. Available at  \href{https://github.com/murphytju/DiffFAS}{https://github.com/murphytju/DiffFAS}.

\end{abstract}

%% file: sec/1_intro.tex
\section{Introduction}

Face recognition~\cite{kim2022adaface} has been developed remarkably and been widely applied in many systems. However, it is vulnerable to physical presentation attacks, such as print attacks \cite{boulkenafet2017oulu,zhang2012face}, replay attacks \cite{chingovska2012effectiveness,wen2015face}, 3D-mask attacks \cite{george2019biometric,rostami2021detection}, and novel types of presentation attacks are continually emerging. Therefore, face anti-spoofing~\cite{atoum2017face,sun2023rethinking,wang2022domain,wang2022patchnet,jia2020single} plays an increasingly important role in face recognition systems.

With the emergence of academic datasets \cite{boulkenafet2017oulu,zhang2012face,chingovska2012effectiveness,wen2015face,george2019biometric,rostami2021detection} over the past decade, deep learning-based FAS \cite{atoum2017face}
has achieved significant performance gains. 
In order to enhance generalization capabilities, researchers have delved into multi-source domain training approaches \cite{shao2019multi,wang2022patchnet,jia2020single,wang2022domain,sun2023rethinking}. However, the utilization of multiple source domains gives rise to domain shifts, manifesting as \textit{image style} and \textit{image quality} challenges. As elucidated in \cite{zou2023adversarial}, image quality is quantified by factors such as image degradation, blur, and low resolution, while image style is characterized by local image statistics \cite{wang2022domain}, reflecting the color\input{subtex/firstpage}and texture differences between images of the same or different types of attacks, as shown in Fig.~\ref{fig:FIRSTPAGE}. Both quality and style introduce erroneous prior information to the network, thereby posing a formidable obstacle in learning precise spoof cues.

Existing domain generalization methods address domain shifts by incorporating human prior information, 
such as domain-specific details \cite{shao2019multi,jia2020single} and ID information \cite{wang2022patchnet}, which should be considered disturbances and need to be disregarded.
However, recognizing the imprecision of these human priors, we propose employing \textit{image quality} as a quantifiable metric \cite{mittal2012no} to directly against domain shift. For the unquantifiable \textit{image style} factors, we advocate for the application of generative methods \cite{ho2020denoising} as a viable solution.

Limited works focus on generative perspective to reduce domain shift impact. The implicit augmentation method~\cite{wang2022domain} performs well on cross-domain print and replay attack. However, as a specific designed module, it is challenging to be intergrating into other FAS method.
Additionally, existing FAS systems are susceptible to novel attack methods \cite{yu2022deep}, but labeled data with novel type of attacks is extremely scarce in current academic datasets \cite{george2019biometric}. 
Therefore, explicit generation should be explored.

Existing generative methods mainly base on GAN \cite{goodfellow2020generative} and VAE \cite{kingma2013auto}.
GAN-based methods, such as STDN \cite{liu2020disentangling}, excel in transferring fine-grained attack textures, but do not address the domain shift between print and replay attacks, rendering them unsuitable for these types of attack's domain generalization. On the other hand, 
VAE-based methods \cite{wu2021dual} are limited by their generative capacities, often leading to local texture distortions which restrict their applicability in RGB-based methods. These approaches tend to overfit to limited data. For example, when fed a face image, the current model generates spoof data that closely resembles the original spoof face in the dataset. This generation method results in the erosion of unique identity features and provides only limited assistance for FAS tasks.
Diffusion models \cite{ho2020denoising} have demonstrated exceptional performance in the realm of high-quality image generation. However, the denoising process, as discussed in \cite{si2023freeu}, tends to suppress high-frequency information, encompassing vital texture cues crucial for FAS tasks. Additionally, existing diffusion models are prone to overfitting with limited data.

In this paper, we address domain shift from separate perspective of image style and image quality. For FAS tasks, live faces can be easily collected, but acquiring spoof faces is time-consuming and labor-intensive \cite{yu2022deep}. Thus, we perform live-spoof transformation to generate large number of spoof samples from live faces, as shown in Fig. \ref{fig:FIRSTPAGE}. To better leverage diffusion models, we define the training objective as the reconstruction process of spoof images, which allows the network directly learning spoof texture to avoid texture degradation while mitigating the impact of overfitting. Additionally, we decouple spoof texture by condition guidance and design Spoofing Style Fusion Module (STFM) to suppress the extraction of identity information from the conditional branch. Finally, to preserve the identity information of live (face) images, we utilize image editing algorithms \cite{valevski2022unitune} to obtain spoof versions of live images. For image quality, we integrate quality quantification scores into loss function, introducing quality prior information during the training of classification model. This guides classification model to deeply mine the spoof cues presented in high-quality images, and directly mitigating the impact of image quality shift. We summarize our contributions as follows:

\begin{itemize}
    \item We represent FAS domain shifts as quantifiable and unquantifiable components, namely image quality shift and image style shift, which provide a new perspective for multi-source domain face anti-spoofing.
    
\end{itemize}

\begin{itemize}
    \item We redesign the training and inference processes to tackle texture degradation and overfitting in diffusion models, introducing STFM for decoupled texture guidance. Our DiffFAS framework delivers high-fidelity, identity-consistent spoof generation, mitigating labeled data scarcity. To our best knowledge, this is the first work to introduce diffusion models for FAS.
    
\end{itemize}

\begin{itemize}
    \item We first introduce quality prior information for DG-FAS, and integrate quantifiable quality scores with the loss function to counter cross-domain image quality shifts.
\end{itemize}

\begin{itemize}
    \item We achieve state-of-the-art performance in the cross-domain and cross-attack FAS protocol. High-fidelity generative sample improve upon the best baseline 4.4$\%$ average ACER in the setting of WMCA unseen protocol, demonstrating the effectiveness of the generative method.
\end{itemize}

%-------------------------------------------------------------------------

%% file: subtex/firstpage.tex
%\begin{figure}{r}{0.5\textwidth}
\begin{figure}[t]
  \centering
  \includegraphics[width=0.85\linewidth]{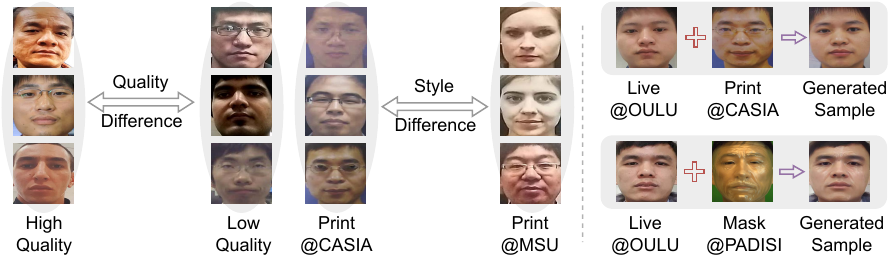}
  %\fbox{\rule{0pt}{2in} \rule{0.9\linewidth}{0pt}}
   %\includegraphics[width=0.8\linewidth]{egfigure.eps}
   \caption{DiffFAS identifies quality and style as separable elements (see the left part) in images and enables cross-domain, cross-attack generation (see the right part) to counteract style discrepancies due to domain shifts.}
   \label{fig:FIRSTPAGE}
   %\vspace{-2.2em}
\end{figure}

%% file: sec/2_Related_work.tex
\section{Related Work}
\label{sec:formatting}

%-------------------------------------------------------------------------
\textbf{Domain Generalized Face Anti-Spoofing:}
Recently, deep learning-based methods \cite{liu2019deep,yang2014learn,kim2019basn,jourabloo2018face} have demonstrated remarkable performance gains but encountered challenges in generalizing to unknown domains, posing hurdles for industrial deployment. Addressing this issue, multi-source domain FAS approaches \cite{shao2019multi,wang2022patchnet,jia2020single,wang2022domain,sun2023rethinking} have garnered increasing attention. SSDG \cite{jia2020single} enhanced network performance by providing attack type information and constraining the network to learn more compact clusters for live samples. PatchNet \cite{wang2022patchnet}, building on SSDG's priors, introduced more precise category labels and employs patch-wise learning to obfuscate identity information. SSAN \cite{wang2022domain} addressed domain shifts' image style factor by employing cross-domain implicit style shuffling. SA-FAS \cite{wang2023domain} advocated for aligning the live-spoof classification hyperplanes across different domains. Different from the aforementioned methods, we approach domain shifts by explicitly considering quality and style, enhancing the model's generalization performance. This is achieved by incorporating diverse human prior inputs specifically developed to counter quality and style shifts.

%-------------------------------------------------------------------------
\noindent \textbf{Generative Method for Face Anti-Spoofing:}
In recent years, there are a few Generative FAS methods \cite{liu2020disentangling,wu2021dual,wang2022domain,wang2023domain}. Wang \etal \cite{wang2023domain} found that simple data augmentation techniques like color transformations are beneficial for training anti-spoofing models. Liu \etal \cite{liu2020disentangling} trained an anti spoofing model by disentangling spoof trace from spoof samples for discrimination and simultaneously adding spoof traces on live samples to enhance classification model's training. Wu \etal \cite{wu2021dual} utilized VAE to learn a joint distribution of identity and spoofing pattern in latent space, thereby generating paired live and spoof images with new identity from random noise. Ho \etal \cite{ho2022classifier} proposed classifier-free guidence, combining condition guidance with the training objective of diffusion model \cite{ho2020denoising}, pioneering the condition diffusion model and leading to diverse generation with multi-modal guidance \cite{zhang2023adding}. 
However, diffusion models face challenges such as texture degradation and overfitting. To leverage the excellent generative capability of diffusion models, we redesign the generative process of diffusion models to produce identity-consistent spoof faces across multiple attack types and styles. This approach effectively addresses domain shift style factors and lessens the need for new attack type labeled data in FAS tasks.

%% file: sec/3_method.tex
\section{Method}

The proposed method, including the DiffFAS generative framework and the Sample-Level Relative Quality loss, which will be introduced next. DiffFAS generative framework consists of the following three critical properties: 1) \textit{Precise spoofing style label}; 2) \textit{Texture preservation}; 3) \textit{Identity-consistency}; Labeled data helps to counter the limitations inherent in academic datasets, with texture being a crucial factor in distinguishing spoof faces. Maintaining identity consistency enables the clear observation of the advantages of authentic spoof generation, without the introduction of new identities. Then, the Relative Quality loss for classification is proposed. This strategy maximizes the extraction of discriminative cues from high-quality (HQ) images in FAS datasets, effectively preventing overfitting to low-quality (LQ) images. 

\input{subtex/framework_fig}
%-------------------------------------------------------------------------
\subsection{Preliminary}

\textbf{Diffusion Model.} Diffusion models \cite{ho2020denoising,nichol2021improved} are generative models that are trained to predict an image from random noise through a gradual denoising process, which consists of the forward process and reverse process. The forward process progressively adds noise to the original image as 
\begin{equation}
    q({y}_{t}|{y}_{t-1}) = \mathcal{N}({y}_{t};\sqrt{1-{\beta }_{t}}{y}_{t-1},{\beta }_{t}I),
\end{equation}
where $t\backsim [1,T]$ and ${\beta }_{1},{\beta }_{2},...,{\beta }_{T}$ is a fixed variance schedule with ${\beta }_{t}\in \left ( 0,1\right )$,
until the original image ${y}_{0}$ is degraded into Gaussian noise ${y}_{t}$. The true posterior $p({y}_{t-1}|{y}_{t})$ can be approximated by training a U-Net model ${\epsilon }_{\theta }$ to predict the initial noise $\epsilon $ with the MSE loss as 
\begin{equation}
{\mathcal{L}}_{mse} = {\mathbb{E}}_{t\backsim [1,T],{y}_{0}\backsim q({y}_{0}),\epsilon}{\left \| \epsilon -{\epsilon }_{\theta }({y}_{t},t)\right \|}^{2}.
\end{equation}
Then in the reverse process, we can sample from a Gaussian noise as 
\begin{equation}
{p}_{\theta }({y}_{t-1}|{y}_{t}) = \mathcal{N}({y}_{t-1};{\epsilon }_{\theta }({y}_{t},t)).
\end{equation}

\noindent\textbf{Generation Objective.} 
We define spoofing style as various novel attack types and distinct styles of the same attack type.
Consider multiple FAS datasets as $\left \{ {D}_{1},{D}_{2},...{D}_{N}\right \}$, with a live face ${X}_{l}^{{id}_{1}}$ from ${D}_{i}$ and a spoof face ${Y}_{s}^{{id}_{2}}$ from ${D}_{j}$. Here, $X$ and $Y$ signify different spoofing styles. Our objective is to generate ${Y}_{s}^{{id}_{1}}$, transforming ${X}_{l}^{{id}_{1}}$ into ${Y}_{s}^{{id}_{1}}$ while adopting the spoofing style of ${Y}_{s}^{{id}_{2}}$ as follows: 
\begin{equation}
{Y}_{s}^{{id}_{1}} = {\epsilon }_{\theta }({Y}_{s}^{{id}_{2}} , {X}_{l}^{{id}_{1}}).
\label{Eq.4}
\end{equation}
However, due to the scarcity of cross-domain data pairs is the training process, we simplify this formula to 
\begin{equation}
{X}_{s}^{{id}_{1}} = {\epsilon }_{\theta }({X}_{s}^{{id}_{2}} , {X}_{l}^{{id}_{1}}),
\end{equation}
where the training exclusively involves same-domain image pairs $({X}_{l}^{{id}_{1}},{X}_{s}^{{id}_{1}})$. The training and inference process is illustrated in Fig.~\ref{fig:framework}.

%-------------------------------------------------------------------------
\subsection{DiffFAS Framework}
\textbf{Overall description.} 
To combat texture degradation and excessive smoothing, DiffFAS merges the training objective with the generative goal through spoof image reconstruction. We use live (face) images from paired sets, matched with randomly selected spoof images of the same style, to direct the reconstruction process of spoof images.
This process is formulated as follows:
\begin{equation}
\label{Eq.6}
{X}_{t} = \sqrt{\bar{{\alpha }_{t}}}{X}_{s}^{{id}_{1}} + \sqrt{1-\bar{{\alpha }_{t}}}\epsilon ,
\end{equation}
\begin{equation}
{p}_{\theta }({X}_{t-1}|{X}_{t}) = \mathcal{N}({X}_{t-1};{\epsilon }_{\theta }(({X}_{t},{X}_{l}^{{id}_{1}}),t,{X}_{s}^{{id}_{2}})),
\end{equation}
where ${\alpha }_{t} = 1 - {\beta }_{t}$, $\bar{{\alpha }_{t}} = \textstyle\prod_{i=1}^{t}{\alpha }_{i}$. To maintain identity consistency, the live image ${X}_{l}^{{id}_{1}}$ is concatenated with the noisy version of spoof image ${X}_{s}^{{id}_{1}}$ and input into U-Net. During training progress, this operation aligns the reconstructed spoof image with the corresponding live image.

\noindent\textbf{Spoofing Style Fusion Module.} To effectively assimilate substantial spoof information from the randomly chosen guide image ${X}_{s}^{{id}_{2}}$, we utilize a pre-trained ResNet \cite{he2016deep} denoted as $\phi $. During the encoder training, precise spoofing style labels are provided, and the network $\phi $ is trained until convergence with cross-entropy loss. As the conditional control branch, feature maps are extracted from the outputs of layers to serve as conditions as
\begin{equation}
{Cond}_{1},{Cond}_{2},...,{Cond}_{n} = \phi ({X}_{s}^{{id}_{2}}).
\end{equation}

Employing feature maps directly as conditions in our model could influence the identity characteristics of the generated image during cross-domain or cross-attack inference scenarios. To effectively extract pertinent spoofing style features from the condition feature maps and concurrently minimizing identity distortion, we design Spoofing Style Fusion Module (STFM). 
Inspired by style transfer methodologies \cite{huang2017arbitrary,jing2020dynamic} and face synthesis works \cite{kim2023dcface}, the means and variance of an image are employed to represent its style. Since we need to handle the fine-grained spoofing style carried by the spoof images, we partition the feature map into small patches, denoted as $\left \{{P}_{1},{P}_{2},...{P}_{N}\right \}$. For each patch, we compute the mean and variance independently, thereby obtaining the representation that is rich in localized information. These computed means and variances are subsequently flattened into the arrays.

In our framework, we propose to deploy asymmetric patch sizes in the means and variance branches. This approach is informed by the observation that, pooling operations, characterized by patch-wise means, tend to preserve identity information. In contrast, the n-neighborhood variance is primarily indicative of texture-related details. Consequently, opting for a larger patch size in the means branch becomes advantageous, as it effectively minimizes the influence of identity-related information. This strategic choice of patch size differentiation is pivotal in ensuring that the focus remains predominantly on the textural aspects, thereby enhancing the fidelity of spoofing style representation while concurrently attenuating identity interference. Then the global means and variance of the feature map are concatenated to incorporate comprehensive global style information into our model. The condition map $\left \{{Cond}_{1},{Cond}_{2},...,{Cond}_{n}\right \}$  is processed into an array of means and variances, represented as $\left \{ {\mu }_{cond},{\sigma }_{cond}\right \}$. 
Parallel to this, analogous operations are executed on the feature maps derived from the U-Net backbone, generating a corresponding set of means and variances $\left \{ {\mu }_{x},{\sigma }_{x}\right \}$. The final stage involves performing cross-attention on these two distinct sets of sequences as follows:
\begin{equation}
{\mu }_{f} = attn(Q({\mu }_{x}),KV({\mu }_{cond})),
\end{equation}
\begin{equation}
{\sigma }_{f} = attn(Q({\sigma }_{x}),KV({\sigma }_{cond})).
\end{equation}

Once the fused feature means and variances are obtained, we remove the last element global information, resize, and upsample them to match the feature map's size with bilinear interpolation. Subsequently, we combine the processed means and variance matrix, yielding the fused spoofing style feature map as
\begin{equation}
{X}_{f}= X + conv(norm(X)*{\sigma }_{f} + {\mu }_{f}).
\end{equation}

Then, we apply STFM to all U-Net blocks with dimensions less than or equal to 32$\times$32.
%-------------------------------------------------------------------------
\subsection{Inference}
\textbf{Disentangled Guidance.}
As the model learns the mapping between image pairs in the dataset, it inherently acquires the capability to perform spoof generation. However, a critical challenge arises in controlling the intensity of the spoof strength. Excessively high spoof strength can inadvertently impact the identity information, leading to deviations from the original ID. To address this issue and enable better control over the spoofing style, we integrate classifier-free guidance \cite{ho2022classifier} into our framework. Specifically, spoofing style condition will be setting to zero with a certain probability. During the sampling phase, the denoising outputs for both conditional and unconditional scenarios are computed\input{subtex/loss_fig}independently at each timestep. By subtracting these outputs, we yield the specific influence of the guidance. Consequently, the sampling process is redefined as follows:
\begin{equation}
\begin{aligned}
\label{Eq.12}
{p}_{\theta }({X}_{t-1}|{X}_{t}) = \mathcal{N}({X}_{t-1};{\epsilon }_{\theta }(({X}_{t},{X}_{l}^{{id}_{1}}),t)) + \\ 
\gamma *(\mathcal{N}({X}_{t-1};{\epsilon }_{\theta }(({X}_{t},{X}_{l}^{{id}_{1}}),t,{X}_{s}^{{id}_{2}}))-\\
\mathcal{N}({X}_{t-1};{\epsilon }_{\theta }(({X}_{t},{X}_{l}^{{id}_{1}}),t))).
\end{aligned}
\end{equation}

The influence of the spoofing style guidance on the inference stage can be controlled by adjusting the parameter $\gamma$. 

\noindent \textbf{Sampling.}
To enhance the model's ability to retain identity information and alleviate the effects of potentially problematic samples in datasets—such as image pairs with mismatched IDs—we draw inspiration from Unitune \cite{valevski2022unitune}, namely, instead of directly sampling noise, our framework employs an image-editing sampling approach. Specifically, we introduce $t$ steps of random noise to the original image, following Eq.~\ref{Eq.6}, and then perform conditional sampling from the noisy live image, following Eq.~\ref{Eq.12}.

%-------------------------------------------------------------------------
\subsection{Sample-Level Relative Quality Loss}

As a primary manifestation of domain shifts, we propose that furnishing the network with quality prior information can significantly improve its performance. Instead of simple binary classification, we classify data based on specific spoofing styles. Recognizing the high similarity within each spoofing style class, we employ margin softmax loss to learn a compact clusters, similar to techniques found in cosface \cite{wang2018cosface} or arcface \cite{deng2019arcface}. These methods introduce an angular margin $\psi$ or additive margin $\omega$ to the classifier, modifying the original softmax formula as:
\begin{equation}
\cos (\theta )\dashrightarrow g(\theta) = \cos (\theta +\psi ) - \omega .
\end{equation}

Inspired by Adaface \cite{kim2022adaface}, 
we realize that network attention towards certain images can be controlled through additive and angular margin. This allows us to integrate inherent image attributes with the loss, such as the image quality\input{subtex/maintab}\input{subtex/WMCA} score BRISQUE \cite{mittal2012no}, as illustrated in Fig.~\ref{fig:loss_equivalent_analysis} (a). Notably, this score is inversely related to image quality. In FAS tasks, LQ images often contain spoof cues associated with local blurriness, which are not ideal for intense mining using margin softmax loss. Therefore, our goal is to focus more on mining from HQ images, while avoiding over-extraction from LQ images. First, we calculate the absolute quality scores ${b}_{aq} = \left \{ {b}_{1},{b}_{2},...,{b}_{N}\right \}$ for a batch, along with the average quality score and variance ${b}_{mean}$, ${b}_{var}$ and the number of samples below or above the average, denoted as ${n}_{1}, {n}_{2}$. Next, we convert the distribution of these quality scores into a standard normal distribution:
\begin{equation}
{b}_{norm} = \frac{{b}_{mean}-{b}_{aq}}{({b}_{std}+eps)}.
\end{equation}

Note that normalized score is directly proportional to image quality. We then scale it according to the 3$\sigma$ rule to ensure that a significant number of samples fall within a certain interval, and then limit the distribution to the range $\left [ {-n}_{1}/{N},{n}_{2}/{N}\right ]$. This range approaches [-0.5,0.5] as the sample capacity tends to infinity, but extremely small scores within a batch can cause the sample mean is lower than the population mean and lead to ${n}_{1} < {n}_{2}$, while a larger ${n}_{2}/{N}$ can increase the margin applied to extreme samples, allowing adaptive\input{subtex/generate_padisi_ocim_fig}handling of such extreme cases. The relative quality scores are calculated as
\begin{equation}
{b}_{rq} = torch.clip(\frac{{b}_{norm}}{3},\frac{-{n}_{1}}{N},\frac{{n}_{2}}{N}).
\end{equation}

This ensures that the means of each batch are centered at zero on the relative quality axis, and each sample's margin is adaptively adjusted based on the distance between ${b}_{mean}$ and their scores. Next, we use the relative quality score to calculate the margin. The loss formula is computed as:
\begin{equation}
\psi = -s*{b}_{rq}, \omega  = s*(1+{b}_{rq}),
\end{equation}
\begin{equation}
 g(\theta) = \cos (\theta - s*{b}_{rq} ) - s*(1+{b}_{rq}),
\end{equation}
\begin{equation}
{L}_{RQ} = -\frac{1}{n}\displaystyle\sum_{i=1}^{n}\log_{}{\frac{{e}^{m*g(\theta)}}{{e}^{m*g(\theta)}+\textstyle\sum_{j=1,j\ne {y}_{i}}^{{c}}{e}^{\cos ({\theta }_{j})} }},
\end{equation}
where $m$ is a scale coefficient. The analysis of the RQ loss is shown in Fig. \ref{fig:loss_equivalent_analysis} (b),
our objective is to mine information from HQ samples. Consequently, as the training advances, the corresponding additive margin for HQ images increases, while for LQ images, it decreases. Taking inspiration from SSDG \cite{jia2020single}, where clusters formed by live and spoof classes exhibit distinct compactness, we introduce different adjustment hyper-parameters denoted as $s$. This ensures that the cluster formed by live images becomes more compact.

%% file: subtex/framework_fig.tex
\begin{figure*}
  \centering
  %\vspace{-1.3em}
    \includegraphics[width=0.9\linewidth]{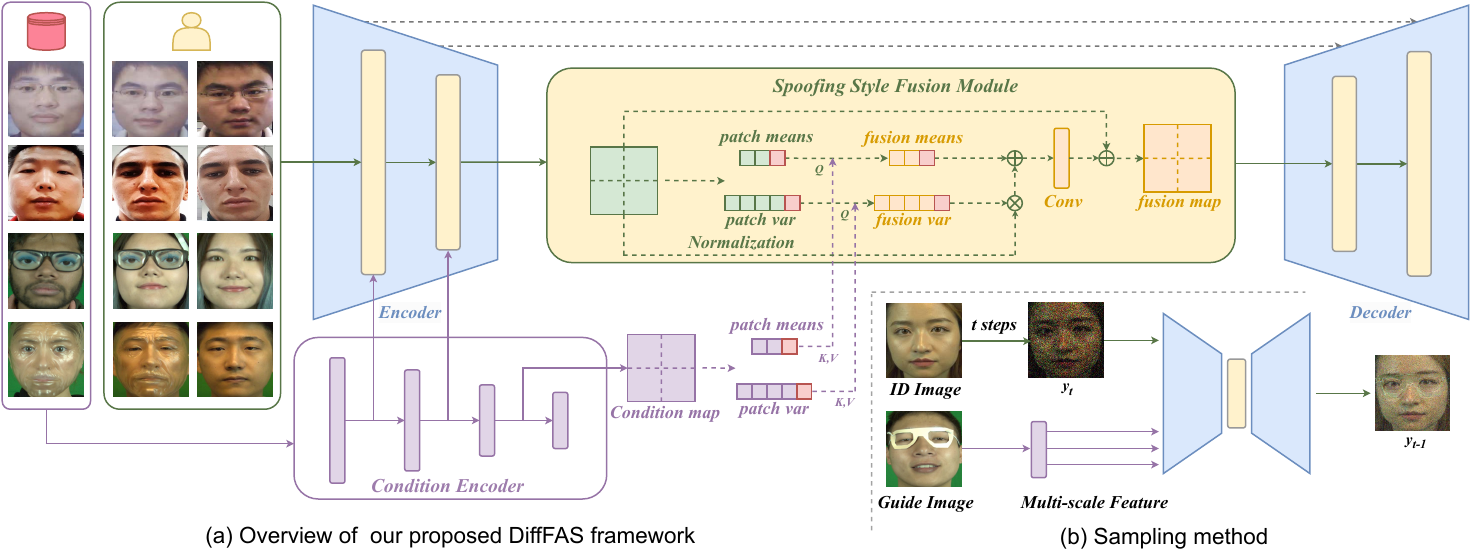}
  %\hfill
  %\vspace{-1.3em}
  \caption{The proposed DiffFAS Generative Framework is a UNet-based network composed of a Spoofing style encoder and a noise prediction module, training with Spoofing Style Pool (the first column) and Live-Spoof Pair Images (the second column). The spoofing style encoder extract texture of the random selected spoof image, and get multi-scale features from different encoder layers. 
  We design an asymmetric Spoofing Style Fusion Module (STFM) to reduce the introduction of identity information of the conditional branch, and achieve information aggregation with the backbone through cross-attention.
  This allows the network to fully capture the spoof texture and achieve high-fidelity spoof synthesis. During the Inference stage, we employ image editing techniques for the sampling process, enhancing our control over the inference stage, and further improving the consistency between the generative ID and the original ID.}
  \label{fig:framework}
  %\vspace{-10pt}
\end{figure*}

%% file: subtex/loss_fig.tex
\begin{figure*}[htbp]
  \begin{minipage}{0.45\textwidth}
    
    \caption{(a) Examples for different domain's live sample BRISQUE score, and lower score means higher quality. (b) Visualization of the equivalent additive margin function $F({b}_{rq}, \theta) = \cos (\theta +\psi ) - \omega -\cos (\theta )$, with different scale coefficient 0.2, 0.4. As training progresses, the equivalent additive margin for HQ samples is larger, while LQ images are conversely.}
    \label{fig:loss_equivalent_analysis}
  \end{minipage}
  \begin{minipage}{0.55\textwidth}
    \centering
    \includegraphics[width=0.7\linewidth]{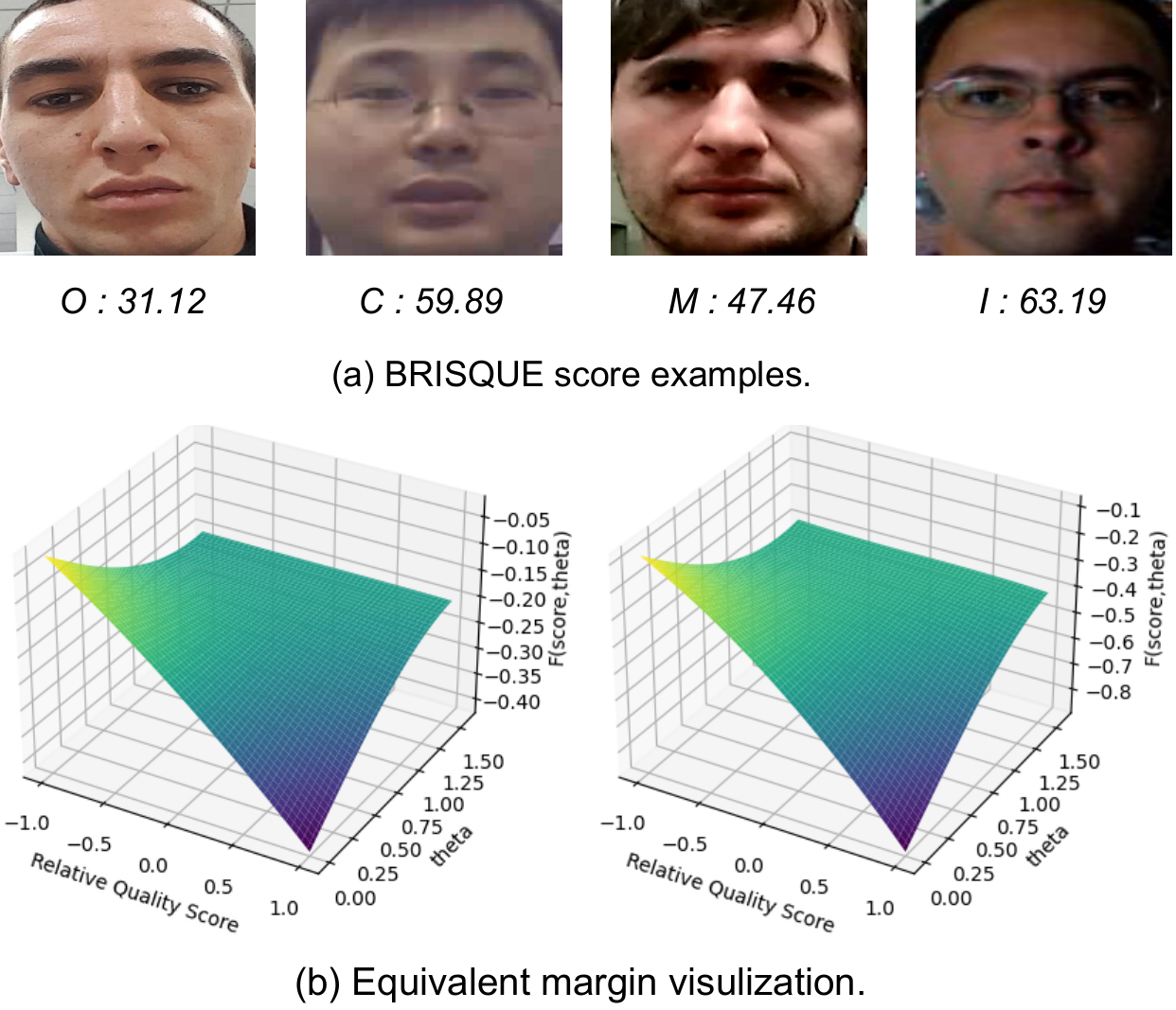}
  \end{minipage}%
  
\end{figure*}

%% file: subtex/maintab.tex
\begin{table*}[t]

%\vspace{-0.2em}
    \caption{
        \small 
        Comparisons with SoTA methods:
        Evaluated on four popular benchmark datasets with leave one out protocol: CASIA (\textbf{C}), Idiap Replay (\textbf{I}), MSU-MFSD (\textbf{M}), and Oulu-NPU (\textbf{O}).  
        $\uparrow$ indicates larger values are better, and $\downarrow$ indicates smaller values are better.
    }
    %\vspace{-0.8em}
    \small\centering
    \scalebox{0.65}{
        \begin{tabular}{rrrrrrrrrrr} \toprule
        \multicolumn{1}{c}{\multirow{2}{*}{\textbf{Method}}} & \multicolumn{2}{c}{\textbf{OCI$\rightarrow$M}} & \multicolumn{2}{c}{\textbf{OMI$\rightarrow$C}} & \multicolumn{2}{c}{\textbf{OCM$\rightarrow$I}} & \multicolumn{2}{c}{\textbf{ICM$\rightarrow$O}}& \multicolumn{2}{c}{\textbf{Average}} \\
        \multicolumn{1}{c}{} & \multicolumn{1}{c}{\textbf{HTER }$\downarrow$} & \multicolumn{1}{c}{\textbf{AUC} $\uparrow$} & \multicolumn{1}{c}{\textbf{HTER  $\downarrow$}} & \multicolumn{1}{c}{\textbf{AUC $\uparrow$}} & \multicolumn{1}{c}{\textbf{HTER}  $\downarrow$} & \multicolumn{1}{c}{\textbf{AUC} $\uparrow$} & \multicolumn{1}{c}{\textbf{HTER}  $\downarrow$} & \multicolumn{1}{c}{\textbf{AUC}  $\uparrow$}& \multicolumn{1}{c}{\textbf{HTER}  $\downarrow$} & \multicolumn{1}{c}{\textbf{AUC}  $\uparrow$}
         \\ \midrule
        MADDG \cite{shao2019multi} & 17.69 & 88.06 & 24.50 & 84.51 & 22.19 & 84.99 & 27.98 & 80.02&23.09&84.36 \\
        SSDG-M \cite{jia2020single} & 16.67 & 90.47 & 23.11 & 85.45 & 18.21 & 94.61 & 25.17 & 81.83 &20.79&88.09\\
        %DiVT-R \cite{liao2023domain} & 11.43 & 94.68 & 18.67 & 91.32 & 21.43 & 88.28 & 17.48 & 89.97&17.25&91.06 \\
        NAS-FAS \cite{yu2020fas} & 19.53 & 88.63 & 16.54 & 90.18 & 14.51 & 93.84 & 13.80 & 93.43&16.09&91.52 \\
        DRDG \cite{liu2021dual} & 12.43 & 95.81 & 19.05 & 88.79 & 15.56 & 91.79 & 15.63 & 91.75&15.67&92.04 \\
        SSAN-M \cite{wang2022domain} & 10.42 & 94.76 & 16.47 & 90.81 & 14.00 & 94.58 & 19.51 & 88.17 &15.10&92.08\\
        SSDG-R \cite{jia2020single} & 7.38 & 97.17 & 10.44 & 95.94 & 11.71 & 96.59 & 15.61 & 91.54&11.29&95.31 \\
        SSAN-R \cite{wang2022domain} & 6.67 & \textbf{98.75} & 10.00 & 96.67 & 8.88 & 96.79 & 13.72 & 93.63&9.82&96.45 \\
        PatchNet \cite{wang2022patchnet} & 7.10 & 98.46 & 11.33 & 94.58 & 13.40 & 95.67 & 11.82 & 95.07&10.91&95.95 \\
        SA-FAS \cite{sun2023rethinking} & 5.95 & 96.55 & 8.78 & 95.37 & 6.58 & 97.54 & \textbf{10.00} & \textbf{96.23} &7.83&96.42\\
        DiffFAS-R (ours) & \textbf{5.90} & 98.10 &\textbf{ 7.32} &\textbf{ 97.40 }& \textbf{5.66}  & \textbf{98.69} & 12.38 & 94.33 &\textbf{7.82}&\textbf{97.13} \\ 
        DiffFAS-V (ours) & \textbf{2.86} & 98.41 & 10.11 & 96.32 & 6.36  & 97.89 & \textbf{8.11} & \textbf{97.27}&\textbf{6.86}&\textbf{97.47} \\ \bottomrule
        \end{tabular}
     }

    \label{tab:best}
    %\vspace{-1ex}
    %\vspace{-0.2em}
\end{table*}

%% file: subtex/WMCA.tex
%\vspace{-0em}
\begin{table*}[t]
 \caption{Results of seen and unseen protocols on WMCA dataset. The values ACER(\%) reported on testing sets are obtained with thresholds computed for BPCER=1\% on development sets. The best results are bolded.}
   %\vspace{-0.8em}
\centering
\setlength{\tabcolsep}{11pt}
  \resizebox{11.9cm}{!}{
        \begin{tabular}{llllllllll}
    \toprule[1pt]
    {\textbf{Method}} & {\textbf{Seen}} & \multicolumn{7}{c}{\textbf{Unseen}}                  & {\textbf{Average}} \\
\cmidrule{3-9}          &       & Glasses & Rigid Mask & Fake Head & Flexible Mask & Paper Mask & Print & Replay &  \\
    \midrule[1pt]
CCL \cite{liu2022contrastive} &27.14&35.13&15.10&21.82&\textbf{7.18}&18.91&20.53&11.79&18.64\\
DiVT-M \cite{liao2023domain}   &6.89&34.52&4.86&10.01&22.56&6.07&2.65&23.11&14.83\\
EPCR \cite{wang2023consistency} & $-$ &\textbf{16.00}&3.40&0.70&49.7&\textbf{0.20}&\textbf{0.10}&\textbf{3.70}&10.50\\
MCDeepPixBiS \cite{george2019deep} & 5.68&24.72&3.29&0.78&24.55&2.14&1.19&20.89&11.08\\
DiffFAS-V              & \textbf{4.25} & 19.62   & \textbf{2.17}      & \textbf{0.00}      & 13.72         & 0.43       & 0.87  & 9.97   & \textbf{6.68}   \\ 
    \bottomrule[1pt]
  \end{tabular}}%
  \label{tab:intra_WMCA}%
 %\vspace{-4em}
\end{table*}

%% file: subtex/generate_padisi_ocim_fig.tex
\begin{figure*}[t]
  % \centering
  \begin{minipage}[b]{0.74\linewidth}
    \includegraphics[width=0.98\linewidth]{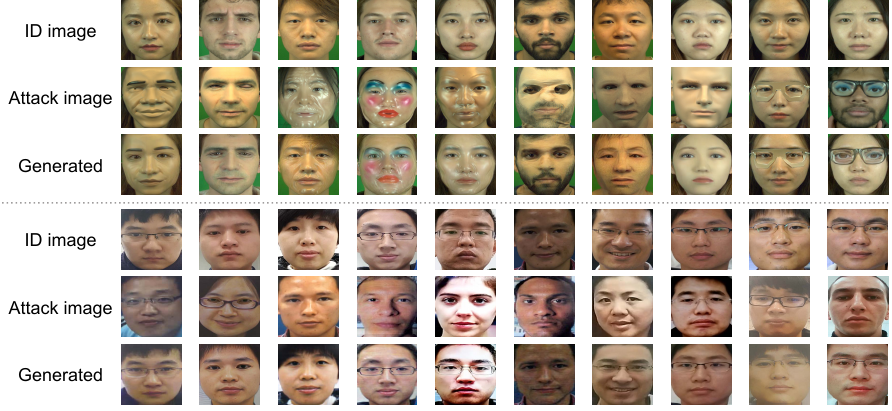}
  \end{minipage}%
  \begin{minipage}[b]{0.26\linewidth}
    \captionof{figure}{Generative samples for cross-attack on PADISI (up) and cross-domain on OCIM (down). The first row is the ID image, and the second row is the randomly selected guide image.}
    \label{fig:padisi}
  \end{minipage}
\end{figure*}

%% file: sec/4_experiment.tex
\section{Experiments}

%-------------------------------------------------------------------------
\subsection{Experimental Setups}
\textbf{Datasets.} We evaluate our framework on six benchmarks, including the widely used domain generalization protocol OULU-NPU (O) \cite{boulkenafet2017oulu}, CASIA-MFSD (C) \cite{zhang2012face}, Idiap Replay Attack (I) \cite{chingovska2012effectiveness}, MSU-MFSD (M) \cite{wen2015face}, containing various styles of print and replay attack; WMCA \cite{george2019biometric} and PADISI \cite{rostami2021detection}, which contains multiple types of novel attacks, such as flexible masks or glasses. 

\noindent \textbf{Metrics.} The model's performance is evaluated using the following standard metrics: Half Total Error Rate (HTER), Area Under Curve (AUC), and Average Classification Error Rate (ACER).

%-------------------------------------------------------------------------
\subsection{Implementation}
\textbf{Generative Experiment.} 
The goal of OCIM is cross-domain generation, specifically involving the generating of spoof samples with the ID (image) from one domain and the spoofing style from another domain. This aims to assess the generative capability of the model. In the case of WMCA and PADISI, numerous live faces lack corresponding spoof faces with specific attack types. Consequently, our objective is to facilitate cross-attack type generation, ensuring the acquisition of all types of attacks for each live face.

\noindent \textbf{Classification Experiment.} This experiment is to  evaluates the impact of proposed methods on classification networks. Here, the input images are cropped with RetinaFace \cite{deng2020retinaface} and resized to 256$\times$256. Based on our experience, MobileViT \cite{mehta2021mobilevit} is used as the backbone. Moreover, for the OCIM experiment, we conduct additional trials using ResNet-18 \cite{he2016deep} to ensure a fair comparison with previous methods. For experiments with less training data, such as WMCA series and ICM→O, we adopt Adam optimizer with learning rate 1e-4, while others with SGD optimizer with learning rate 2e-3. In cross domains or attacks experiments, we only utilize data from visible domains or attacks for generation, ensuring that data leakage does not occur.

\input{subtex/exp_56}\input{subtex/exp_34}

%-------------------------------------------------------------------------
\subsection{Results of Generative Experiment}

%\vspace{1.5em}
The generative samples of cross-attack and cross-domain settings are presented in Fig.~\ref{fig:padisi}. For PADISI's various types of mask attacks, our method can learn specific texture features on masks with high homogeneity in the training set, and replace skin texture with attack texture while maintaining ID consistency. As such, it can enrich the diversity of masks in the dataset. For PADISI face-edit attacks, our method can perform partial replacement of the image edited part, without modifying the sampling algorithm. Note that in PADISI, our model demonstrates robust texture learning capabilities despite the constraint of having a limited number of image pairs (20-30) for training. In the case of OCIM's domain shift, we achieve cross-domain style transfer by leveraging universal texture features, thereby underscoring the versatility and generality inherent in our texture-based approach.

%-------------------------------------------------------------------------

%\subsection{Performance}
\subsection{Results of Classification Experiment}
\textbf{Cross-Domain Performance.} Table~\ref{tab:best} summarizes our OCIM cross-domain performance compared with an extensive collection of recent studies. DiffFAS achieves state-of-the-art performance on all protocols. 
Because of the scarcity of HQ images in ICM→O and considering that ResNet has a comparatively weaker information extraction capability compared to MobileViT, DiffFAS-R shows diminished effectiveness when compared to other protocols.
The results of cross-domain testings on WMCA and PADISI are presented in Table~\ref{tab:wmca-padisi}, which contains several types of novel attacks. HTER$\%$ values reports on testing sets are obtained with the fixed threshold of 0.5 on testing sets. Our method shows greater benefits under testing protocols with limited data.

\noindent \textbf{Cross-attack Performance.} The results of cross-attack experiments on WMCA datasets are presented in Table~\ref{tab:intra_WMCA}, comprising the grandtest protocol and seven unseen tests for seven attack types. Leveraging our remarkable mask generation capability, DiffFAS attains a state-of-the-art performance in both seen and average unseen attack types.

\noindent \textbf{Limited Source Domains Performance.} Furthermore, we evaluate the performance of our method in cases of limited availability of source domains. Specifically, M and I are selected as the source domains for training while the remaining two are used as the target domains for testing, respectively. As shown in Table~\ref{tab:limitedsource}, our method achieves the lowest HTER and the highest AUC despite limited source data, which demonstrates the modeling efficiency and generalization capability of our model.

%% file: subtex/exp_56.tex
\noindent
%\vspace{-1em}
\begin{minipage}[t]{.48\linewidth}    
%\begin{table}[t]
    
    %\vspace{-0.8em}
  \captionof{table}{Comparison on limited source domains.}

    %\caption{Comparison on limited source domains.} 
    %\vspace{-0.8em}
    \small\centering
        \resizebox{5.2cm}{!}{
        \begin{tabular}{lllll}\toprule
        \multicolumn{1}{c}{\multirow{2}{*}{\textbf{Method}}} & \multicolumn{2}{c}{\textbf{MI$\rightarrow$C}} & \multicolumn{2}{c}{\textbf{MI$\rightarrow$O}}  \\
        \multicolumn{1}{c}{} & \multicolumn{1}{c}{\textbf{HTER }$\downarrow$} & \multicolumn{1}{c}{\textbf{AUC} $\uparrow$} & \multicolumn{1}{c}{\textbf{HTER  $\downarrow$}} & \multicolumn{1}{c}{\textbf{AUC $\uparrow$}} 
         \\ \midrule
        SSDG-R \cite{jia2020single} &19.86 & 86.46 & 27.92 & 78.72 \\
        SSAN-R \cite{wang2022domain} &25.56 & 83.89 & 24.44 & 82.56 \\
        DiVT-M \cite{liao2023domain} &20.11 & 86.71 & 23.61 & 85.73 \\
        DiffFAS-V &\textbf{15.06} & \textbf{92.83} & \textbf{16.19} & \textbf{92.62} \\ \bottomrule
        \end{tabular}    }
        
    \label{tab:limitedsource}
    %\vspace{-1.5ex}
 %\end{table}   
\end{minipage}%
\hfill
\begin{minipage}[t]{.48\linewidth}

   \captionof{table}{Comparison on WMCA, PADISI datasets.}
    %\caption{Comparison on WMCA, PADISI datasets.}   
    \small\centering
        \resizebox{6cm}{!}{
        \begin{tabular}{lllll}\toprule
        \multicolumn{1}{c}{\multirow{2}{*}{\textbf{Method}}} & \multicolumn{2}{c}{\textbf{WMCA$\rightarrow$PADISI}} & \multicolumn{2}{c}{\textbf{PADISI$\rightarrow$WMCA}}  \\
        \multicolumn{1}{c}{} & \multicolumn{1}{c}{\textbf{HTER }$\downarrow$} & \multicolumn{1}{c}{\textbf{AUC} $\uparrow$} & \multicolumn{1}{c}{\textbf{HTER  $\downarrow$}} & \multicolumn{1}{c}{\textbf{AUC $\uparrow$}} 
         \\ \midrule
        MCDeepPixBiS \cite{george2019deep} &12.83 & 95.43 & 17.18 & 90.96   \\ 
        EPCR \cite{wang2023consistency} &10.06 & 97.30 & 16.26& 92.49   \\ 
         
        DiffFAS-V w/o generation &8.67 & 97.06 & 15.15 & 92.56   \\ 
        DiffFAS-V w/o loss &8.13 & 97.59 & 14.96 & 93.19   \\
        DiffFAS-V &\textbf{7.09} & \textbf{97.92} & \textbf{12.67} &\textbf{94.04} \\ \bottomrule
        \end{tabular} }   
        
    \label{tab:wmca-padisi}
   % \end{table}
%\end{wraptable}
% Your second table here
\end{minipage}

%% file: subtex/exp_34.tex
\noindent
\begin{minipage}[t]{.48\linewidth}    
%\begin{table}[t]
    
    %\vspace{-0.8em}
  \captionof{table}{Comparisons with previous loss function.}

    \small\centering
        \resizebox{6cm}{!}{
        \begin{tabular}{lll} \toprule
        \multicolumn{1}{c}{\multirow{2}{*}{\textbf{Method}}} & \multicolumn{2}{c}{\textbf{OCI$\rightarrow$M}}   \\
        \multicolumn{1}{c}{} & \multicolumn{1}{c}{\textbf{HTER }$\downarrow$} & \multicolumn{1}{c}{\textbf{AUC} $\uparrow$} 
         \\ \midrule
        Resnet18  &7.71 & 98.05   \\ 
        Resnet18 + Focal loss \cite{lin2017focal}  &11.48 & 94.28 \\
        Resnet18 + Asymmetric softmax loss \cite{wang2022patchnet}  &6.81 & 97.64 \\
         Resnet18 + Single-side triplet loss \cite{jia2020single}  &6.57 & 97.90 \\
       Resnet18 + RQ loss &5.90 & 98.10 \\\bottomrule
        \end{tabular}    }        
    
    \label{tab:otherloss}
    %\vspace{-1.5ex}
 %\end{table}   
\end{minipage}%
\hfill
\begin{minipage}[t]{.48\linewidth}

   \captionof{table}{Ablation of RQ loss live/spoof scaling factors.}
    \small\centering
        \resizebox{4.6cm}{!}{
        \begin{tabular}{lll} \toprule
        \multicolumn{1}{c}{\multirow{2}{*}{\textbf{Scaling parameter}}} & \multicolumn{2}{c}{\textbf{OCI$\rightarrow$M}}   \\
        \multicolumn{1}{c}{} & \multicolumn{1}{c}{\textbf{HTER }$\downarrow$} & \multicolumn{1}{c}{\textbf{AUC} $\uparrow$} 
         \\ \midrule
         0.4/0.1  &6.57 & 98.31   \\ 
         0.4/0.2 &3.10 & 99.63 \\
         0.4/0.3 &4.23 & 99.09 \\
         0.4/0.4 &4.86 & 98.51 \\\bottomrule
        \end{tabular}    }        
    \label{tab:SCALE-HYPERPARAMETER}
   % \end{table}
%\end{wraptable}
% Your second table here
\end{minipage}

%% file: sec/5_ablation.tex
\input{subtex/exp_910}

%-------------------------------------------------------------------------

\section{Ablation Study}

\subsection{Generation Method Ablation}
\textbf{Ablation of STFM.} We establish a basic style transfer module, employing means and variance substitution of the backbone and condition as the baseline, comparing with baseline incorporating patch-wise mean and variance, STFM without patch, and proposed STFM. 
Generative samples are presented in Fig.~\ref{fig:w/o_module}.\input{subtex/w_o_module}\input{subtex/exp_12}It is evident that, even with the editing algorithm in the baseline method, preserving the original identity information poses a challenge, resulting in distortion for specific attack types. The inclusion of patch information significantly improves the mitigation of distortion. Furthermore, it is observed that employing a larger mean patch aids in minimizing the impact on identity information while introducing attack texture to live faces.

\noindent \textbf{Ablation of Inference Hyper-parameter.} The strength of the spoof feature can be adjusted by manipulating the forward process timesteps $t$ and CFG parameters $\gamma$. However, adding excessively strong spoof traces may potentially compromise the identity information. Ablation experiments on these two parameters are illustrated in Fig.~\ref{fig:forwardtimesteps} and~\ref{fig:spoofstyleparam}. It is evident that as $t$ increases or $\gamma$ increases, the spoof feature become more significant. This capability allows us to create samples with varying levels of spoof strength, thereby enhancing the robustness of the classification network.

\noindent \textbf{Comparison with existing works in preventing overfitting.} In this section, we conduct comparative experiment to observe DiffFAS's capability against overfitting, which is a common challenge encountered by previous generative methods. Earlier methods suffer from overfitting due to data scarcity, leading to generative spoof images bearing a strong resemblance to the samples in the original dataset. The reduction in generative diversity diminishes the utility of synthetic datasets for classification tasks, thereby undermining the intended objective of generation. To validate our method's capability against overfitting, we conduct comparative experiments with DSDG \cite{liu2021dual}. We convert the live samples generated by DSDG into spoof samples with the same attack type, enabling us to observe the generative differences between above two methods, and results are show in Fig. \ref{fig:compare2dsdg}. It is apparent that DSDG faces challenges in learning the joint distribution between live and spoof samples, which can be attributed to the extremely small scale of data in PADISI.

\input{subtex/compare2dsdg}\input{subtex/DSDG}

\subsection{Classification Method Ablation}
\textbf{Effectiveness of Framework Components.}  
The Table~\ref{tab:wmca-padisi} and~\ref{tab:MI-CO_ablation} presents both generative samples and Relative Quality loss gain for classification experiment. In different scenarios, the two major factors of domain shift: quality and style, play distinct yet complementary roles. Existing experiments confirm that these components effectively counteract the dual characteristics of domain shift, successfully mitigating quality and style shifts, thereby enhancing generalization performance.
Regarding the RQ loss, we utilize different scaling factors for live and spoof samples, and the ablation study on these factors is provided in Table~\ref{tab:SCALE-HYPERPARAMETER}.

\noindent \textbf{Comparison with Previous Works.} 
To facilitate a fair comparison of the impact of our method with previous generative approaches on cross-domain FAS tasks, we establish a baseline without generative data. This baseline involves binary classification with CrossEntropy loss. Subsequently, we introduce DSDG and DiffFAS data into the experiment, and the results are presented in Table~\ref{tab:compare_todsdg}. Additionally, to validate the effectiveness of the proposed loss, we replace various previously proposed SOTA losses with DiffFAS-R, including Focal loss \cite{lin2017focal}, Asymmetric softmax loss from PatchNet \cite{wang2022patchnet}, and the Single-side triplet loss from SSDG \cite{jia2020single}, with CrossEntropy loss as the baseline. The results are shown in Table~\ref{tab:otherloss}. Notably, with our generative samples, SSDG and PatchNet demonstrate improved performance, highlighting the effectiveness of our generative method.

%% file: subtex/exp_910.tex
\noindent
%\vspace{-1em}
\begin{minipage}[t]{.48\linewidth}    
%\begin{table}[t]
    
    %\vspace{-0.8em}
   \captionof{table}{Ablation of Framework Components.}
    %\caption{Comparison on WMCA, PADISI datasets.}   
   \small\centering
        \resizebox{6cm}{!}{
        \begin{tabular}{lllll}\toprule
        \multicolumn{1}{c}{\multirow{2}{*}{\textbf{Method}}} & \multicolumn{2}{c}{\textbf{MI$\rightarrow$C}} & \multicolumn{2}{c}{\textbf{MI$\rightarrow$O}}  \\
        \multicolumn{1}{c}{} & \multicolumn{1}{c}{\textbf{HTER }$\downarrow$} & \multicolumn{1}{c}{\textbf{AUC} $\uparrow$} & \multicolumn{1}{c}{\textbf{HTER  $\downarrow$}} & \multicolumn{1}{c}{\textbf{AUC $\uparrow$}} 
         \\ \midrule
        MobileVit &24.57 & 83.70 & 24.16 & 85.37   \\
        DiffFAS-V w/o loss &16.91 & 91.82 & 18.02 & 89.98   \\
        DiffFAS-V w/o generation &15.50 & 92.53 & 16.99 & 91.19   \\
        DiffFAS-V &15.06 & 92.83 & 16.19 & 92.62 \\ \bottomrule
        \end{tabular}    }
        
    \label{tab:MI-CO_ablation}
    %\vspace{-1.5ex}
 %\end{table}   
\end{minipage}%
\hfill
\begin{minipage}[t]{.48\linewidth}

     \captionof{table}{Comparisons with previous generative methods.}

    %\caption{Comparison on limited source domains.} 
    %\vspace{-0.8em}
    \small\centering
        \resizebox{4.7cm}{!}{
        \begin{tabular}{lll} \toprule
        \multicolumn{1}{c}{\multirow{2}{*}{\textbf{Method}}} & \multicolumn{2}{c}{\textbf{OCI$\rightarrow$M}}   \\
        \multicolumn{1}{c}{} & \multicolumn{1}{c}{\textbf{HTER }$\downarrow$} & \multicolumn{1}{c}{\textbf{AUC} $\uparrow$} 
         \\ \midrule
         Resnet18  &9.67 & 95.96   \\ 
         Resnet18 + DSDG \cite{liu2021dual} &8.57 & 96.55 \\
         Resnet18 + DiffFAS &7.71 & 98.05 \\\bottomrule
        \end{tabular}    }
    
    \label{tab:comparetodsdg}
   % \end{table}
%\end{wraptable}
% Your second table here
\end{minipage}

%% file: subtex/w_o_module.tex
\begin{figure*}[t]
  \centering
  \begin{minipage}[b]{0.75\textwidth}
    %\centering
    \includegraphics[width=0.95\linewidth]{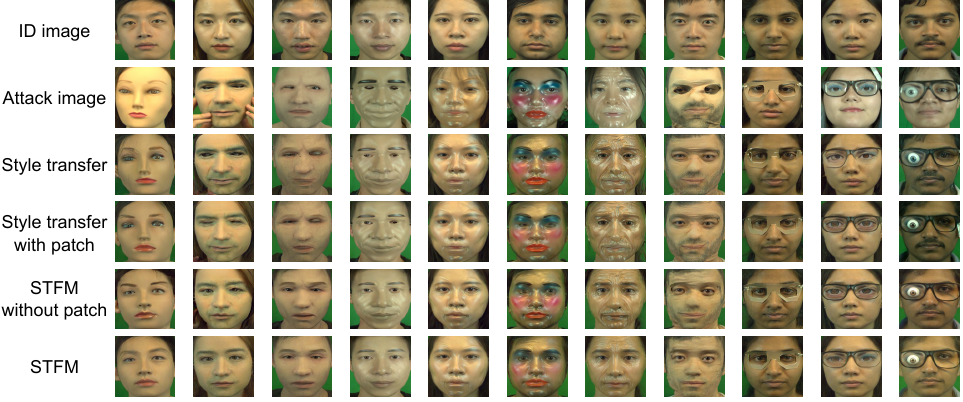}
  \end{minipage}%
  \begin{minipage}[b]{0.25\textwidth}
    \captionof{figure}{Visualization with various baselines. It is evident that DiffFAS achieves the balance between identity and spoof texture.}
    \label{fig:w/o_module}
  \end{minipage}
\end{figure*}

%% file: subtex/exp_12.tex
\noindent
\begin{minipage}[t]{.49\linewidth}
    \centering
    \includegraphics[width=0.8\linewidth]{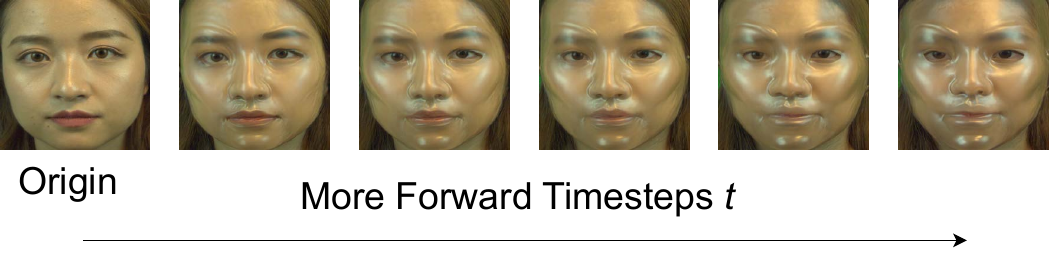}
    \captionof{figure}{Ablation of the impact of time steps: with $\gamma$ = 2.0, $t$ = 100, 300, 500, 700, 900 from left to right.}
    \label{fig:forwardtimesteps}
\end{minipage}%
\hfill
\begin{minipage}[t]{.49\linewidth}
    \centering
    \includegraphics[width=0.8\linewidth]{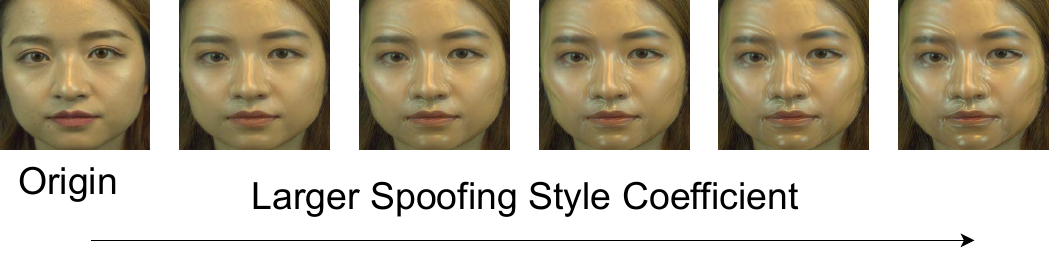}
    \captionof{figure}{Ablation of $\gamma$ impact: with $t$ = 100, $\gamma$ = 1.2, 1.5, 2.0, 2.5, 3.0 from the left to the right.}
    \label{fig:spoofstyleparam}
\end{minipage}

%% file: subtex/compare2dsdg.tex
\begin{figure*}[t]
  \centering
  \begin{minipage}[b]{0.6\textwidth}
    \centering
    \includegraphics[width=0.95\linewidth]{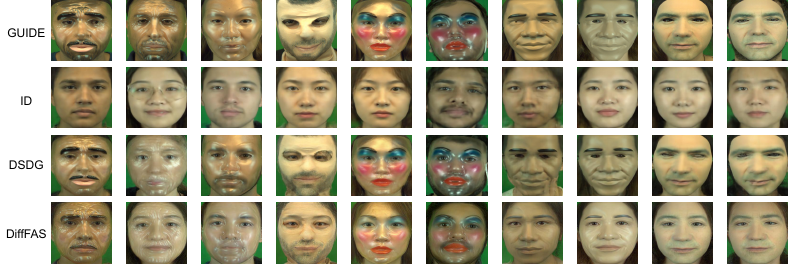}
  \end{minipage}%
  \hfill
  \begin{minipage}[b]{0.4\textwidth}
    \captionof{figure}{Comparison with DSDG on PADISI: The ID image is sourced from DSDG's generated live samples to ensure a fair competition.}
    \label{fig:compare2dsdg}
  \end{minipage}
\end{figure*}

%% file: subtex/DSDG.tex
\begin{wraptable}{r}{0.3\textwidth}
    %\vspace{-1.5em}
    \caption{Comparisons with previous generative methods.}     
     
    \small\centering
        \resizebox{3.5cm}{!}{
        \begin{tabular}{lll} \toprule
        \multicolumn{1}{c}{\multirow{2}{*}{\textbf{Method}}} & \multicolumn{2}{c}{\textbf{OCI$\rightarrow$M}}   \\
        \multicolumn{1}{c}{} & \multicolumn{1}{c}{\textbf{HTER }$\downarrow$} & \multicolumn{1}{c}{\textbf{AUC} $\uparrow$} 
         \\ \midrule
         Resnet18  &9.67 & 95.96   \\ 
         Resnet18 + DSDG \cite{liu2021dual} &8.57 & 96.55 \\
         Resnet18 + DiffFAS &7.71 & 98.05 \\\bottomrule
        \end{tabular}    }
    
    \label{tab:compare_todsdg}
    %\vspace{-1.0ex}
    %\vspace{-0.8em}
\end{wraptable}

%% file: sec/6_conclusion.tex
%-------------------------------------------------------------------------

\section{Conclusion}

This paper proposes a novel approach to address domain shifts by separating them into measurable quality and style components. We introduce the Relative Quality loss that integrates quality scores into the training loss function, adding a quality prior to the classification model. Further, we present DiffFAS, a versatile generative model for high-fidelity cross-domain and cross-attack generation, addressing the lack of labeled data for novel attack types.
Our model currently faces a constraint in open-world face reconstruction, requiring corresponding ID in the training set to generate its spoof version. Enhancing this capability is a key focus for future research. Investigating methods to broaden the model's capacity for realistic spoof generation, independent of specific training pairs, holds promise for further development.

%% file: sec/X_suppl.tex
\clearpage
\setcounter{page}{1}
%\maketitlesupplementary

%\section{More Experiment Results}
\noindent In the main paper, we introduce DiffFAS, a novel diffusion-based generative method designed to reduce cross-domain style shifts. Additionally, we propose the Relative Quality (RQ) loss function, which utilizes quality priors to address cross-domain quality shifts. The supplementary material includes further analysis of these methods through a series of experiments.

\section{Ablation Study about STFM}

In this section, we present an ablation study evaluating the effects of different patch sizes for both means and variance patches within the Spoofing Style Fusion Module (STFM). Main paper investigates the relationship between image statistics and some image factors, including color, shape, and texture. 
To assess the influence of these statistics on the generative spoof samples, we conducted experiments to evaluate the impact of different patch sizes in STFM across various attack scenarios, including mask attacks in the PADISI dataset and print/replay attacks in the OCIM dataset. The results of experiments with means as shown in Fig. \ref{fig:patchsize} (a), revealing that excessively small means patches lead to significant alterations in facial features, such as replacing by the mask ID or blending with the guide image's identity information. 
On the contrary, larger variance patches might overlook details in the generative samples, such as the color smears on the surface of plastic masks and punched eye holes in printed attacks, as illustrated in Fig. \ref{fig:patchsize} (b). Additionally, an increase in the size of the variance patches leads to a reduction in the intensity of the attack texture. Therefore, we recommend the means patch size of 6$\times$6 and variance patch size of 2$\times$2 to achieve a balanced representation of identity and spoof information in the generative samples.

In contrast, DiffFAS utilizes an image editing-based sampling algorithm that effectively integrates identity information from live faces while closely matching the dataset's spoof texture patterns, thus leveraging the overfitting to datasets' texture feature and preserving the model's generative diversity. 
Furthermore, we assess the objective quality of the generative samples by calculating the BRISQUE scores for both the generative and original datasets. These scores are obtained by averaging the BRISQUE scores across all samples in each dataset, and comparison results are shown in Fig. \ref{fig:score}. It is observed that in the PADISI dataset, the generative quality of DiffFAS slightly surpasses that of the original dataset and considerably outperforms the generative samples of DSDG \cite{liu2021dual}. For the OCIM dataset, the average quality score of DiffFAS falls between those of the high-quality and low-quality datasets, modestly outperforming DSDG.

\section{Results on More Datasets}
In this section, we present additional generative results of the WMCA dataset, which includes various illumination conditions, backgrounds, and head pose conditions, contrasting with the uniform image conditions in the PADISI dataset. The results are shown in Fig. \ref{fig:wmca} (a). Experiments conducted with the WMCA\input{subtex/sec2}\input{subtex/score}\input{subtex/hter}dataset highlight the adaptability of our method in handling complex data environments. To further observe the effectiveness of our method in countering overfitting, we conduct generative experiment with the image pairs employed during the training process, assessing the differences between the generative samples and ground-truth. Results are illustrated in Fig. \ref{fig:wmca} (b). It is observed that the generative samples retain live face's identity information, avoiding the excessive resemblance to the ground-truth. The result validate the efficacy of our proposed generative process, significantly mitigating\input{subtex/tsne}\input{subtex/weight_sampling}the overfitting and improving generative diversity. Additionally, we also conducted the most challenging cross-domain and cross-attack experiments on the OULU-PADISI datasets, as illustrated in Fig. \ref{fig:wmca} (c). The outstanding generative results enable us to extensively generate attack samples using existing facial datasets, thereby significantly reducing the cost of anti-spoofing dataset collection.

\section{Effectiveness of Quality Prior}
To verify the effectiveness of the quality prior, we implement weighted sampling method as a simple baseline on ResNet18, OCI-M protocal, and results are shown in Tab. \ref{tab:weight_sampling}. For OULU dataset, the probability of random sampling ranges between threefold and fourfold. Increasing the sampling probability of high-quality samples can achieve certain performance gains, but overweighting high-quality samples leads to overfitting and performance degradation. To further observe the effectiveness of RQ loss, we conduct experiments on the OCIM protocol, and utilize t-SNE \cite{van2008visualizing} to observe the sample distribution of a live and spoof class during the training process. The results are illustrated in the Fig. \ref{fig:tsne}, where the color of point representing the BRISQUE score of the sample, as denoted by the color bar on the right. During the training process, a classification boundary progressively emerges between the live and spoof classes, evolving from stage (a) to (b). Subsequently, high-quality data gradually move towards the class center, as depicted in stage (c), ultimately resulting in the distribution observed in stage (d). This phenomena confirms the effectiveness of the proposed RQ loss in integrating quality priors into the network. Additionally, we apply the proposed RQ loss in several SOTA methods to validates the compatibility of quality prior, such as SSDG \cite{jia2020single} and PatchNet \cite{wang2022patchnet}. The results of the OCI→M experiment are shown in Fig. \ref{fig:hter}, demonstrating that the integration of quality prior information can work in conjunction with the prior information proposed in previous methods, such as SSDG's adversarial prior and PatchNet's patch-wise learning, ultimately enhancing the performance of these methods.

\input{subtex/sec1}

%% file: subtex/sec2.tex
\begin{figure*}[t]
  \centering
  \includegraphics[width=1.0\linewidth]{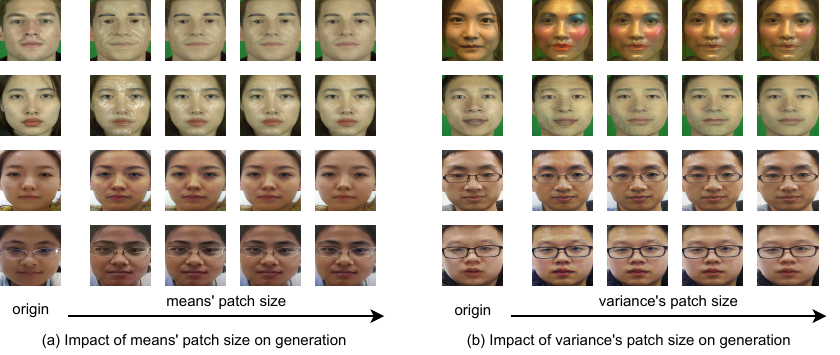}
  %\fbox{\rule{0pt}{2in} \rule{0.9\linewidth}{0pt}}
   %\includegraphics[width=0.8\linewidth]{egfigure.eps}

   \caption{Ablation result of means and variance's patch size impact on generative samples, implemented on PADISI and OCIM datasets. From left to right, the size of the patches gradually increases.}
   \label{fig:patchsize}
   %\vspace{-1em}
   
\end{figure*}

%% file: subtex/score.tex
\begin{figure}[t]
  \centering
  
  \includegraphics[width=1.0\linewidth]{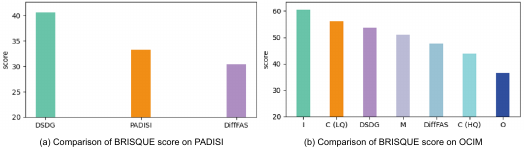}

   \caption{The BRISQUE \cite{mittal2012no} scores of the generative samples from DSDG and DiffFAS on both PADISI (a) and OCIM (b), which reflects the fundamental image quality aspects such as the degree of image degradation, blurriness, and low resolution.}
   \label{fig:score}
   %\vspace{-3em}
\end{figure}

%% file: subtex/hter.tex
\begin{wrapfigure}{l}{0.45\textwidth}
  \centering
  \includegraphics[width=1.0\linewidth]{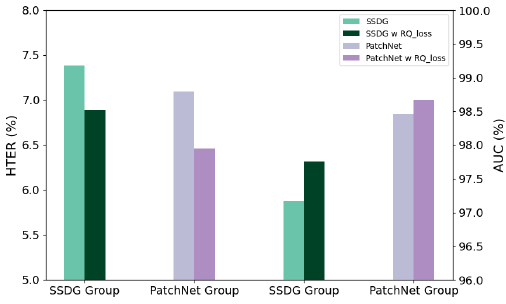}
  %\fbox{\rule{0pt}{2in} \rule{0.9\linewidth}{0pt}}
   %\includegraphics[width=0.8\linewidth]{egfigure.eps}

   \caption{Effects of Implementing RQ Loss on Previous works.}
   \label{fig:hter}
   %\vspace{-1.0em}
\end{wrapfigure}

%% file: subtex/tsne.tex
\begin{figure*}[t]
  \centering
  \includegraphics[width=1.0\linewidth]{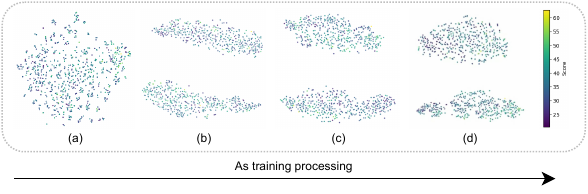}
  %\fbox{\rule{0pt}{2in} \rule{0.9\linewidth}{0pt}}
   %\includegraphics[width=0.8\linewidth]{egfigure.eps}

   \caption{Visualization of the live and spoof samples' distribution during the training process. The color of point represents the BRISQUE score of the sample, as indicated by the color bar on the right, where a lower score signifies higher quality. Figures (a) to (d) represent various stages of training. Figure (a) illustrates that in the initial stages of training, the samples have not yet formed distinct clusters. Figure (b) shows that, as training progresses, the live and spoof samples begin to form initial clusters. Figure (c) depicts samples with lower scores gradually move towards the center of their respective clusters, resulting in the distribution observed in figure (d).}
   \label{fig:tsne}
 
\end{figure*}

%% file: subtex/weight_sampling.tex
\begin{wraptable}{r}{0.5\textwidth}
    %\vspace{-1.5em}
    \caption{2x, 3x, and 4x denote the sampling weights applied to OULU, relative to average sampling.}     
     
    \small\centering
        \resizebox{5cm}{!}{
        \begin{tabular}{lll} \toprule
        \multicolumn{1}{c}{\multirow{2}{*}{\textbf{Experimental setup}}} & \multicolumn{2}{c}{\textbf{OCI$\rightarrow$M}}   \\
        \multicolumn{1}{c}{} & \multicolumn{1}{c}{\textbf{HTER }$\downarrow$} & \multicolumn{1}{c}{\textbf{AUC} $\uparrow$} 
         \\ \midrule
         Average Sampling & 10.67 & 95.11 \\
        2x Rate OULU & 8.76 & 96.25 \\ 
        3x Rate OULU & 8.29 & 96.77 \\
        Random Sampling & 9.67 & 95.96 \\
        4x Rate OULU & 12.86 & 93.84 \\\bottomrule
        \end{tabular}    }
    
    \label{tab:weight_sampling}
    %\vspace{-1.0ex}
    %\vspace{-0.8em}
\end{wraptable}

%% file: subtex/sec1.tex
\begin{figure*}[t]
  \centering
  \includegraphics[width=1.0\linewidth]{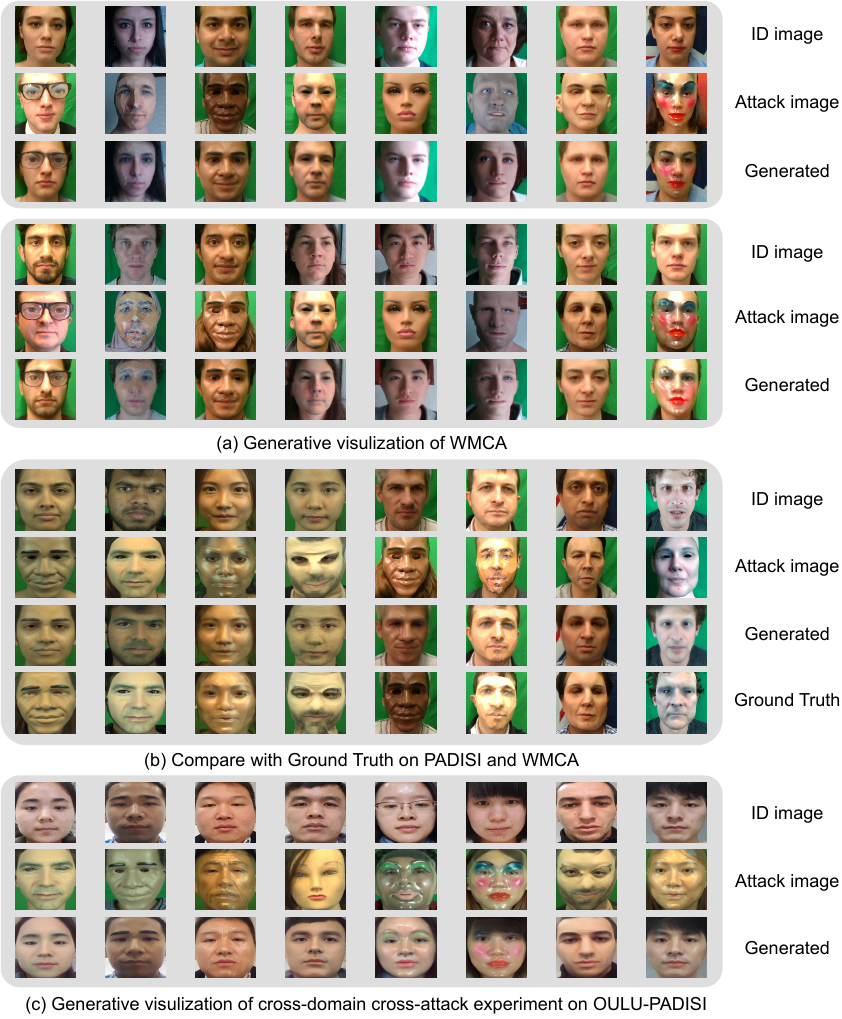}
  %\fbox{\rule{0pt}{2in} \rule{0.9\linewidth}{0pt}}
   %\includegraphics[width=0.8\linewidth]{egfigure.eps}
   %\vspace{-1.8em}
   %\vspace{-1ex}
   \caption{(a) Generative samples for cross-attack experiment on WMCA, including more complex lighting, background, and head poses conditions. (b) Visualization of generative samples for image pairs in the training set, which reveal that our method's capability against overfitting. (c) Generative samples for cross-attack and cross-domain experiments on OULU-PADISI.}
   \label{fig:wmca}
  % \vspace{-0.8em}
   
\end{figure*}

%% file: main.bbl
\begin{thebibliography}{10}
\providecommand{\url}[1]{\texttt{#1}}
\providecommand{\urlprefix}{URL }
\providecommand{\doi}[1]{https://doi.org/#1}

\bibitem{atoum2017face}
Atoum, Y., Liu, Y., Jourabloo, A., Liu, X.: Face anti-spoofing using patch and depth-based cnns. In: 2017 IEEE International Joint Conference on Biometrics (IJCB). pp. 319--328. IEEE (2017)

\bibitem{boulkenafet2017oulu}
Boulkenafet, Z., Komulainen, J., Li, L., Feng, X., Hadid, A.: Oulu-npu: A mobile face presentation attack database with real-world variations. In: 2017 12th IEEE international conference on automatic face \& gesture recognition (FG 2017). pp. 612--618. IEEE (2017)

\bibitem{chingovska2012effectiveness}
Chingovska, I., Anjos, A., Marcel, S.: On the effectiveness of local binary patterns in face anti-spoofing. In: 2012 BIOSIG-proceedings of the international conference of biometrics special interest group (BIOSIG). pp.~1--7. IEEE (2012)

\bibitem{deng2020retinaface}
Deng, J., Guo, J., Ververas, E., Kotsia, I., Zafeiriou, S.: Retinaface: Single-shot multi-level face localisation in the wild. In: Proceedings of the IEEE/CVF conference on computer vision and pattern recognition. pp. 5203--5212 (2020)

\bibitem{deng2019arcface}
Deng, J., Guo, J., Xue, N., Zafeiriou, S.: Arcface: Additive angular margin loss for deep face recognition. In: Proceedings of the IEEE/CVF conference on computer vision and pattern recognition. pp. 4690--4699 (2019)

\bibitem{george2019deep}
George, A., Marcel, S.: Deep pixel-wise binary supervision for face presentation attack detection. In: 2019 International Conference on Biometrics (ICB). pp.~1--8. IEEE (2019)

\bibitem{george2019biometric}
George, A., Mostaani, Z., Geissenbuhler, D., Nikisins, O., Anjos, A., Marcel, S.: Biometric face presentation attack detection with multi-channel convolutional neural network. IEEE transactions on information forensics and security  \textbf{15},  42--55 (2019)

\bibitem{goodfellow2020generative}
Goodfellow, I., Pouget-Abadie, J., Mirza, M., Xu, B., Warde-Farley, D., Ozair, S., Courville, A., Bengio, Y.: Generative adversarial networks. Communications of the ACM  \textbf{63}(11),  139--144 (2020)

\bibitem{he2016deep}
He, K., Zhang, X., Ren, S., Sun, J.: Deep residual learning for image recognition. In: Proceedings of the IEEE conference on computer vision and pattern recognition. pp. 770--778 (2016)

\bibitem{ho2020denoising}
Ho, J., Jain, A., Abbeel, P.: Denoising diffusion probabilistic models. Advances in neural information processing systems  \textbf{33},  6840--6851 (2020)

\bibitem{ho2022classifier}
Ho, J., Salimans, T.: Classifier-free diffusion guidance. arXiv preprint arXiv:2207.12598  (2022)

\bibitem{huang2017arbitrary}
Huang, X., Belongie, S.: Arbitrary style transfer in real-time with adaptive instance normalization. In: Proceedings of the IEEE international conference on computer vision. pp. 1501--1510 (2017)

\bibitem{jia2020single}
Jia, Y., Zhang, J., Shan, S., Chen, X.: Single-side domain generalization for face anti-spoofing. In: Proceedings of the IEEE/CVF Conference on Computer Vision and Pattern Recognition. pp. 8484--8493 (2020)

\bibitem{jing2020dynamic}
Jing, Y., Liu, X., Ding, Y., Wang, X., Ding, E., Song, M., Wen, S.: Dynamic instance normalization for arbitrary style transfer. In: Proceedings of the AAAI conference on artificial intelligence. vol.~34, pp. 4369--4376 (2020)

\bibitem{jourabloo2018face}
Jourabloo, A., Liu, Y., Liu, X.: Face de-spoofing: Anti-spoofing via noise modeling. In: Proceedings of the European conference on computer vision (ECCV). pp. 290--306 (2018)

\bibitem{kim2022adaface}
Kim, M., Jain, A.K., Liu, X.: Adaface: Quality adaptive margin for face recognition. In: Proceedings of the IEEE/CVF conference on computer vision and pattern recognition. pp. 18750--18759 (2022)

\bibitem{kim2023dcface}
Kim, M., Liu, F., Jain, A., Liu, X.: Dcface: Synthetic face generation with dual condition diffusion model. In: Proceedings of the IEEE/CVF Conference on Computer Vision and Pattern Recognition. pp. 12715--12725 (2023)

\bibitem{kim2019basn}
Kim, T., Kim, Y., Kim, I., Kim, D.: Basn: Enriching feature representation using bipartite auxiliary supervisions for face anti-spoofing. In: Proceedings of the IEEE/CVF International Conference on Computer Vision Workshops. pp.~0--0 (2019)

\bibitem{kingma2013auto}
Kingma, D.P., Welling, M.: Auto-encoding variational bayes. arXiv preprint arXiv:1312.6114  (2013)

\bibitem{liao2023domain}
Liao, C.H., Chen, W.C., Liu, H.T., Yeh, Y.R., Hu, M.C., Chen, C.S.: Domain invariant vision transformer learning for face anti-spoofing. In: Proceedings of the IEEE/CVF Winter Conference on Applications of Computer Vision. pp. 6098--6107 (2023)

\bibitem{lin2017focal}
Lin, T.Y., Goyal, P., Girshick, R., He, K., Doll{\'a}r, P.: Focal loss for dense object detection. In: Proceedings of the IEEE international conference on computer vision. pp. 2980--2988 (2017)

\bibitem{liu2022contrastive}
Liu, A., Zhao, C., Yu, Z., Wan, J., Su, A., Liu, X., Tan, Z., Escalera, S., Xing, J., Liang, Y., et~al.: Contrastive context-aware learning for 3d high-fidelity mask face presentation attack detection. IEEE Transactions on Information Forensics and Security  \textbf{17},  2497--2507 (2022)

\bibitem{liu2021dual}
Liu, S., Zhang, K.Y., Yao, T., Sheng, K., Ding, S., Tai, Y., Li, J., Xie, Y., Ma, L.: Dual reweighting domain generalization for face presentation attack detection. arXiv preprint arXiv:2106.16128  (2021)

\bibitem{liu2019deep}
Liu, Y., Stehouwer, J., Jourabloo, A., Liu, X.: Deep tree learning for zero-shot face anti-spoofing. In: Proceedings of the IEEE/CVF Conference on Computer Vision and Pattern Recognition. pp. 4680--4689 (2019)

\bibitem{liu2020disentangling}
Liu, Y., Stehouwer, J., Liu, X.: On disentangling spoof trace for generic face anti-spoofing. In: Computer Vision--ECCV 2020: 16th European Conference, Glasgow, UK, August 23--28, 2020, Proceedings, Part XVIII 16. pp. 406--422. Springer (2020)

\bibitem{mehta2021mobilevit}
Mehta, S., Rastegari, M.: Mobilevit: light-weight, general-purpose, and mobile-friendly vision transformer. arXiv preprint arXiv:2110.02178  (2021)

\bibitem{mittal2012no}
Mittal, A., Moorthy, A.K., Bovik, A.C.: No-reference image quality assessment in the spatial domain. IEEE Transactions on image processing  \textbf{21}(12),  4695--4708 (2012)

\bibitem{nichol2021improved}
Nichol, A.Q., Dhariwal, P.: Improved denoising diffusion probabilistic models. In: International Conference on Machine Learning. pp. 8162--8171. PMLR (2021)

\bibitem{rostami2021detection}
Rostami, M., Spinoulas, L., Hussein, M., Mathai, J., Abd-Almageed, W.: Detection and continual learning of novel face presentation attacks. In: Proceedings of the IEEE/CVF international conference on computer vision. pp. 14851--14860 (2021)

\bibitem{shao2019multi}
Shao, R., Lan, X., Li, J., Yuen, P.C.: Multi-adversarial discriminative deep domain generalization for face presentation attack detection. In: Proceedings of the IEEE/CVF conference on computer vision and pattern recognition. pp. 10023--10031 (2019)

\bibitem{si2023freeu}
Si, C., Huang, Z., Jiang, Y., Liu, Z.: Freeu: Free lunch in diffusion u-net. arXiv preprint arXiv:2309.11497  (2023)

\bibitem{sun2023rethinking}
Sun, Y., Liu, Y., Liu, X., Li, Y., Chu, W.S.: Rethinking domain generalization for face anti-spoofing: Separability and alignment. In: Proceedings of the IEEE/CVF Conference on Computer Vision and Pattern Recognition. pp. 24563--24574 (2023)

\bibitem{valevski2022unitune}
Valevski, D., Kalman, M., Matias, Y., Leviathan, Y.: Unitune: Text-driven image editing by fine tuning an image generation model on a single image. arXiv preprint arXiv:2210.09477  (2022)

\bibitem{wang2022patchnet}
Wang, C.Y., Lu, Y.D., Yang, S.T., Lai, S.H.: Patchnet: A simple face anti-spoofing framework via fine-grained patch recognition. In: Proceedings of the IEEE/CVF Conference on Computer Vision and Pattern Recognition. pp. 20281--20290 (2022)

\bibitem{wang2018cosface}
Wang, H., Wang, Y., Zhou, Z., Ji, X., Gong, D., Zhou, J., Li, Z., Liu, W.: Cosface: Large margin cosine loss for deep face recognition. In: Proceedings of the IEEE conference on computer vision and pattern recognition. pp. 5265--5274 (2018)

\bibitem{wang2023domain}
Wang, W., Liu, P., Zheng, H., Ying, R., Wen, F.: Domain generalization for face anti-spoofing via negative data augmentation. IEEE Transactions on Information Forensics and Security  (2023)

\bibitem{wang2023consistency}
Wang, Z., Yu, Z., Wang, X., Qin, Y., Li, J., Zhao, C., Liu, X., Lei, Z.: Consistency regularization for deep face anti-spoofing. IEEE Transactions on Information Forensics and Security  \textbf{18},  1127--1140 (2023)

\bibitem{wang2022domain}
Wang, Z., Wang, Z., Yu, Z., Deng, W., Li, J., Gao, T., Wang, Z.: Domain generalization via shuffled style assembly for face anti-spoofing. In: Proceedings of the IEEE/CVF Conference on Computer Vision and Pattern Recognition. pp. 4123--4133 (2022)

\bibitem{wen2015face}
Wen, D., Han, H., Jain, A.K.: Face spoof detection with image distortion analysis. IEEE Transactions on Information Forensics and Security  \textbf{10}(4),  746--761 (2015)

\bibitem{wu2021dual}
Wu, H., Zeng, D., Hu, Y., Shi, H., Mei, T.: Dual spoof disentanglement generation for face anti-spoofing with depth uncertainty learning. IEEE Transactions on Circuits and Systems for Video Technology  \textbf{32}(7),  4626--4638 (2021)

\bibitem{yang2014learn}
Yang, J., Lei, Z., Li, S.Z.: Learn convolutional neural network for face anti-spoofing. arXiv preprint arXiv:1408.5601  (2014)

\bibitem{yu2022deep}
Yu, Z., Qin, Y., Li, X., Zhao, C., Lei, Z., Zhao, G.: Deep learning for face anti-spoofing: A survey. IEEE transactions on pattern analysis and machine intelligence  \textbf{45}(5),  5609--5631 (2022)

\bibitem{yu2020fas}
Yu, Z., Wan, J., Qin, Y., Li, X., Li, S.Z., Zhao, G.: Nas-fas: Static-dynamic central difference network search for face anti-spoofing. IEEE transactions on pattern analysis and machine intelligence  \textbf{43}(9),  3005--3023 (2020)

\bibitem{zhang2023adding}
Zhang, L., Rao, A., Agrawala, M.: Adding conditional control to text-to-image diffusion models. In: Proceedings of the IEEE/CVF International Conference on Computer Vision. pp. 3836--3847 (2023)

\bibitem{zhang2012face}
Zhang, Z., Yan, J., Liu, S., Lei, Z., Yi, D., Li, S.Z.: A face antispoofing database with diverse attacks. In: 2012 5th IAPR international conference on Biometrics (ICB). pp. 26--31. IEEE (2012)

\bibitem{zou2023adversarial}
Zou, Z., Wang, Z., Zhang, B., Xu, Y., Liu, Y., Wu, L., Guo, Z., He, Z.: Adversarial domain generalization for surveillance face anti-spoofing. In: Proceedings of the IEEE/CVF Conference on Computer Vision and Pattern Recognition. pp. 6351--6359 (2023)

\end{thebibliography}
